\pdfoutput=1
\documentclass[11pt,a4paper]{article}
\usepackage[hyperref]{acl2021}
\usepackage{times}
\usepackage{latexsym}

\usepackage{microtype}

\aclfinalcopy 

\title{HLE-UPC at SemEval-2021 Task 5: Multi-Depth DistilBERT for Toxic Spans Detection}

\author{Rafel Palliser Sans \and Albert Rial Farràs\\
  Facultat d'Informàtica de Barcelona (FIB)\\
  Universitat Politècnica de Catalunya (UPC - BarcelonaTECH)\\
  Barcelona, Spain\\
  \texttt{\{rafel.palliser, albert.rial\}@estudiantat.upc.edu}
}

\date{}

\begin{document}
\maketitle
\begin{abstract}

This paper presents our submission to SemEval-2021 Task 5: Toxic Spans Detection. The purpose of this task is to detect the spans that make a text toxic, which is a complex labour for several reasons. Firstly, because of the intrinsic subjectivity of toxicity, and secondly, due to toxicity not always coming from single words like insults or offends, but sometimes from whole expressions formed by words that may not be toxic individually. Following this idea of focusing on both single words and multi-word expressions, we study the impact of using a multi-depth DistilBERT model, which uses embeddings from different layers to estimate the final per-token toxicity. Our quantitative results show that using information from multiple depths boosts the performance of the model. Finally, we also analyze our best model qualitatively.

\end{abstract}

\section{Introduction}
SemEval-2021 Task 5: Toxic Spans Detection \citep{pav2020semeval} consists in detecting which spans make a text toxic. This is quite relevant for nowadays lifestyle in which, aggravated by the COVID-19 pandemic, online conversations have become key to communicate with our family, friends and job mates, or socialize through social networks and streaming chats. Being able to moderate all this digital content is crucial in order to promote healthy online conversations and discussions.

To tackle this problem, in HLE-UPC we have used a BERT-based model with a fully-connected layer on top to perform Named-Entity Recognition and Classification (NERC), with the goal of tagging each word as either toxic or not. Moreover, we have studied and proved that the use of information from different-depth layers enriches the final classification.

Our contributions to Toxic Spans Detection are:
\begin{itemize}
    \item The proposal of an ensemble of three different multi-depth DistilBERTs, achieving an F1-score of 68.54\% and being ranked 14th out of 91 teams in the challenge, just 2.29\% below the best performing model.
    \item The study of multi-depth BERT-based models in the task of Toxic Spans Detection, showing an improvement on the performance compared to non-multi-depth architectures.
    \item A qualitative analysis presenting some ethical concerns regarding racial bias.
\end{itemize}

The source code for our model and pipeline is available at \url{https://github.com/rafelps/HLE-UPC-SemEval-2021-ToxicSpansDetection}.

\section{Related work}
\paragraph{Toxicity}
The task in which we are participating is not the first one to focus on text toxicity. Without going any farther, in last year's edition of SemEval we can find Task 12, also known as OffensEval 2020 \cite{zampieri2020semeval2020}, in which the goal was to identify offensive language in multilingual social media data. In the previous year's competition, SemEval 2019, Task 6 \cite{zampieri2019semeval2019} was also tackling the identification and categorization of offensive language in social media.

Some of the models that solved these tasks involve Convolutional Neural Networks (CNN) \cite{mahata-etal-2019-midas}, Long Short Term Memory Networks (LSTM) \cite{pham-hong-chokshi-2020-pgsg} or attention based models \cite{liu-etal-2019-nuli,wiedemann-etal-2020-uhh} branched from the BERT family \cite{devlin2019bert}.

\paragraph{NERC}
All the mentioned models approach the task as Sequence Classification, this is, encoding a whole sentence and providing a unique prediction for it. Toxic Spans Detection, however, goes a step further by asking participants to detect toxic spans, the exact characters or words that make a text toxic. For this reason, instead of modelling the task as sentiment analysis or document/comment classification, it seems more natural to approach it as token classification, generating an output for each token. More specifically, this task could be seen as a Named Entity Recognition and Classification (NERC) task, in which the goal would be to output the most probable sequence of labels (toxic or not) given an input sentence.

In the field of NERC, we also find some interesting models. The state-of-the-art today are attention-based models, usually stemming from transformers such as BERT \cite{devlin2019bert}, which can be easily converted into a token classifier by adding a simple linear layer on top of the per-token output. However, we can also find some other attention-based models using CNNs \cite{baevski2019clozedriven} or even recurrent architectures such as \citet{jiang-etal-2019-improved, strakova-etal-2019-neural, peters2018deep}, which mix BiLSTMs, CNNs or CRF layers.

\section{Data and Methodology}
\label{sec:method}
\subsection{Data Description}
For this task, the organizers provide us with the Toxic Spans Detection (TSD) dataset, also presented in \citet{pav2020semeval}, containing phrases and comments that may contain toxic spans. Together with each comment, there is the set of indices of the characters that are considered toxic.

The TSD dataset is split into three subsets: trial, train and test sets with approximately 700, 8000 and 2000 comments respectively. All the models presented in this work have been trained exclusively on the TSD training set, while the trial set has been used to validate our systems. Finally, the test set has served to evaluate the performance of our final models using the available limited submissions for the competition.

The TSD dataset contains very diverse comments. Some of them seem quite simple, but others may be ambiguous, require context knowledge or an understanding of tone, which makes the task extremely challenging. There are also some words that have been written in an ingenious way, to avoid naïve toxic detectors, or that are bleeped or censored. Following we present a couple of examples, where toxic characters are underlined:
\begin{itemize}
    \item This is a \underline{stupid} example, so thank you for nothing \underline{a!@\#!@}.
    \item \underline{I bet you can't wait to see him behind bars}.
\end{itemize}

\subsection{Data Cleaning}
With a simple data exploration, it can be seen that approximately 90\% of the toxic spans exactly match with word boundaries, but in the remaining cases we find strange cases such as the following ones:

\begin{enumerate}
    \item \textbf{You are an\underline{ idiot}}: There is a whitespace as a toxic span boundary.
    \item \textbf{You \underline{a}re an idiot}: A random singleton character is marked as toxic.
    \item \textbf{Y\underline{ou are an idiot}}: ``Y'' is not marked as toxic but ``ou'' is.
\end{enumerate}

The majority of these inconsistencies are already known by the organizers of the task and other participants. However, they should still be tackled to provide the best data possible to our models. For this reason, we have cleaned the data using three simple steps and following the idea of toxicity coming from complete words but not from single characters. For each group of consecutive annotated toxic offsets:

\begin{enumerate}
    \item Iteratively remove the first or last toxic offset if it belongs to a whitespace. This solves the first type of inconsistencies.
    \item Remove the toxic offset if it is a singleton: a single consecutive character marked as toxic. This helps in the second type of strange cases.
    \item Iteratively left-expand the range of toxic offsets if the previous character is alphanumeric (so it belongs to the same word). Same for right-expansion. This solves the third problem by including the offsets of the whole word as toxic whenever more than one character is marked as so.\footnote{The opposite strategy, discarding words if not all their characters were marked as toxic was also studied but rejected as performed poorer.}
\end{enumerate}

After cleaning the data, almost the totality of the annotations matches word boundaries. On one side this confirms our hypothesis that toxicity comes from words or expressions but not from characters. On the other side, this enables a word by word analysis in a consistent and robust manner. Nevertheless, the task remains challenging given the subjectivity of the annotations.

\subsection{Preprocessing}
Once data is cleaned and before feeding it to the models, we lower case the text and tokenize it using WordPiece \cite{wu2016googles}, the tokenizer used by BERT-based models, which splits text into (usually) sub-word units. Each of these units has its associated token embedding at the first layer of the respective models.

In this step, we also use the information of the already-cleaned toxic offsets to create a per-token binary label regarding its toxicity.

\subsection{Models}

\paragraph{LSTM}
Long Short-Term Memory was introduced in 1991 by \citet{lstm} as an extension of recurrent neural networks (RNNs), providing them with the ability to capture and memorize long-term dependencies and therefore help prevent the vanishing/exploding problems \cite{bengio, pascanu2013difficulty}.

We use an LSTM tagger as our baseline model to determine the lower bound performance that we should compare with. We use it as a first approach to solve the task, even though we know that the sentences of the dataset might be too long for the network to memorize and capture all long-term dependencies and the entire sentence context. As input for this model, we use pre-trained word embeddings from GloVe \cite{glove}.

\paragraph{Attention-based models}
In 2018, Google Research released Bidirectional Encoder Representation from Transformer (BERT) \cite{devlin2019bert} which achieved many state-of-the-art results on different NLP tasks. This success led to the creation of a lot of new models and improvements based on the BERT architecture: DistilBERT, RoBERTa, ALBERT, ...  This architecture uses the same multi-head transformer structure presented by \citet{vaswani2017attention}, which is basically composed of several stacked Transformer blocks/encoders, including self-attention and feed-forward modules. These help the model obtain richer word representations by finding correlations with other tokens in the sentence.

For our task, we use two BERT-based models, BERT and DistilBERT, with a token classification head --a linear layer on top of the hidden state output of the last Transformer encoder--. These models are pre-trained on huge corpus from different sources and fine-tuned for our downstream task.

\paragraph{Multi-depth models}
Based on the previously presented BERT-like models, we implement a modification that consists in feeding the classification layer an augmented embedding for each token. This augmented embedding is formed by concatenating the hidden outputs of different Transformer blocks, instead of using the last output directly as done in common models for token classification. The empirical results show that using embeddings from different layers provides better representations and boosts the model's performance.

\subsection{Postprocessing}
Once a model outputs its predictions, we loop through them and, for those tokens predicted as toxic, we take their offsets and add them to the final set of toxic spans for that sentence.

Additionally, we add a postprocessing step to increase the correctness of our predictions regarding white characters. These are not returned as tokens by the tokenizer but occupy a character offset. For this reason, for each pair of consecutive tokens predicted as toxic, we also include to the final set the offsets of any white characters in between.

\section{Results}
All the results presented in this section have been calculated using the official metric, the F1-score on the predicted toxic offsets. For detailed information please refer to \citet{pav2020semeval}.

\subsection{Model Comparison}
In Table~\ref{tab:results} we report the results for the best configuration of each of our models both in the official trial and test sets. In these results, we can first note that all the models clearly outperform our baseline. Moreover, using the information of multiple layers is proved to be beneficial for this task, as it improves each of the respective base models, by 0.64\% - 1.42\%. Finally, note that although BERT is larger and more powerful than DistilBERT, it performs poorer in the test set. This might be due to the fact that we select our hyperparameters based on the F1-score on the trial set, which is relatively small and may not be representative of the test data. For this reason, our submitted model is a multi-depth DistilBERT, as it provides better generalization within this task and data.

\begin{table*}[ht]
\centering
\begin{tabular}{lcc}
\hline \textbf{Model} & \textbf{F1-score (trial)} & \textbf{F1-score (test)} \\ \hline
LSTM (baseline) & 61.36\% & 62.06\% \\
DistilBERT & 69.04\% & 67.43\% \\
BERT & 69.22\% & 66.45\%\\
Multi-depth DistilBERT & 69.68\% & \textbf{68.22\%}\\
Multi-depth BERT & \textbf{70.01\%} & 67.87\%\\
\hline
\end{tabular}
\caption{\label{tab:results} Performance comparison for various architectures in the official trial (used as validation) and test sets.}
\end{table*}

\subsection{Layer Selection}
In this study, we have trained a multi-depth DistilBERT using the outputs of different layers or transformer blocks to study its impact on the model's performance.

Table~\ref{tab:layers} shows the results for different experiments in which we have concatenated the outputs of the last N layers of DistilBERT before feeding these enlarged hidden states to the fully connected layer that performs classification.

Results show that performance can be improved by adding different block's outputs, but can also degrade when using too many. For DistilBERT, which has 6 transformer blocks, the sweet spot seems to be using the last 3 layers. Using all 6 also provides good results, which may imply that the first layer's output is also quite informative for this task in which words themselves already help predicting their toxicity.

\begin{table}[h]
\centering
\begin{tabular}{cc}
\hline \textbf{Last N layers} & \textbf{F1-score (trial)}\\ \hline
1 & 69.04\% \\
2 & 69.48\% \\
3 & \textbf{69.68\%} \\
4 & 69.11\% \\
5 & 68.94\% \\
6 & 69.48\% \\
\hline
\end{tabular}

\caption{\label{tab:layers} Performance comparison for multi-depth \mbox{DistilBERT} in the trial set using the concatenation of the last N layer's outputs for the final classification.}
\end{table}

\subsection{Ablation Study}
\label{sec:ablation}
In these experiments, we took apart one component of our system at a time to see its effect on the system's performance. The main components of our method are presented in Section~\ref{sec:method}, and details about our implementation can be found in Appendix~\ref{app:imp}.

Table~\ref{tab:ablation} shows the results for this study, in which we can easily see that all components work towards the performance of our model. Apart from the multi-depth component, which has already been studied, Dropout has been key for our giant model to generalize and prevent overfitting the small data. Using Label Smoothing has also helped, letting the model adapt to the intrinsic subjectivity of the annotations.

Regarding data preparation, it can be seen that the cleaning step has been crucial for the good performance of our system, supporting the known quote ``Garbage in, garbage out''. Finally, our simple postprocessing stage has also provided some tenths to the final performance.

\begin{table}[ht]
\centering
\begin{tabular}{lc}
\hline \textbf{Model} & \textbf{F1-score (trial)}\\ \hline
Multi-depth DistilBERT & \textbf{69.68\%} \\
(ours) -- Multi-depth & 69.04\% \\
(ours) -- Dropout & 68.25\% \\
(ours) -- Label Smoothing & 69.17\% \\
(ours) -- Data Cleaning & 66.44\% \\
(ours) -- Postprocessing & 69.38\% \\
\hline
\end{tabular}
\caption{\label{tab:ablation} Ablation study on the system's components. `--' means leaving that component out. Results for the official trial set.}
\end{table}

\subsection{Ensemble}
Given the results we obtained with single models, we found it interesting to mix some of them to see if they were focusing on different parts of data and could improve the predictions while working together.

Following this idea, we created a simple majority-voting ensemble using the multi-depth models with ``last N layers'' for $N=1,3,6$; this is, a base DistilBERT, a model that concatenates the output of the last 3 transformer blocks and another one that uses all 6 layers of DistilBERT.

The final result for this ensemble is 69.34\% in the trial set --used as validation-- and \textbf{68.54\%} in the test set, our best submission. Note that although being worse than our best single model in the trial set, it has better generalization skills and boosts the performance in the unseen test set.

\subsection{Qualitative}
Apart from the quantitative analysis done before, we analyze in a qualitative manner the performance and behaviour of our best model, to see how well detects offensive and toxic words and in which cases it fails.

Below we present some examples of sentences in the dataset together with their ground truth spans and the detection done by the model. The \underline{ground truth} toxic words appear underlined while the \textcolor{red}{prediction} is shown in red.

\paragraph{Correct predictions}
We observe how our system is highly capable of identifying toxic and offensive words, both when they appear alone and in multi-word expressions.

\begin{itemize}
    \item Billy, are you a complete \underline{\textcolor{red}{idiot}}, being thick headed or just not reading what people...
    \item People insist on being \underline{\textcolor{red}{dumb}}. No other explanation.
    \item Could you please \underline{\textcolor{red}{kill yourself}}?
\end{itemize}

\paragraph{Wrong predictions}
However, our system also fails in some challenging comments. As seen below with the word ``poorly'', our method misses some words marked as toxic which are not very offensive or disrespectful but can become toxic due to the context.
\begin{itemize}
    \item People don't buy that \underline{poorly} built Russian houses...
\end{itemize}

In other cases, our system identifies toxicity when it is not annotated, although under our perspective the prediction seems correct. This could be due to the ambiguity of the task or inconsistencies in the annotations. An example of it is the expression ``freaking donkeys'':
\begin{itemize}
    \item These \textcolor{red}{freaking donkeys} all need to be removed from office. I'm so sick and tired of...
\end{itemize}

Finally, our model fails to detect connectors such as ``of'' and ``and'' in between toxic words. In the dataset there are several annotation philosophies: some annotations tend to mark entire expressions as toxic and some others are more word-oriented, excluding connectors between words.
\begin{itemize}
    \item Are these some of those Russian \underline{\textcolor{red}{pieces} of \textcolor{red}{crap}} that they seem to be building all over Alaska.
\end{itemize}

\paragraph{Ethical concerns}
While doing the qualitative analysis we found several examples indicating that there could be racial bias in the predictions of our model, and although it is beyond the scope of the challenge, we found important to pay attention to it. For this reason, we took some examples from the trial set containing comments about races and changed the words referring to races or origin by others. Below we show an example. The first comment belongs to the competition dataset, while the other is a modification of it, with words ``black'' changed for ``white'' and vice versa, and ``Mexican'' changed for ``American''.

\begin{itemize}
    \item \textcolor{red}{Black} folks built this nation and got \underline{\textcolor{red}{lynching}} for the work. \underline{Heck, white folks} can be so mean that when they lost their slaves they invited \textcolor{red}{illegal Mexican immigrants} to do the work black slaves use to do.
    
    \item White folks built this nation and got \underline{\textcolor{red}{lynching}} for the work. \underline{Heck, \textcolor{red}{black} folks} can be so mean that when they lost their slaves they invited \textcolor{red}{illegal American} immigrants to do the work black slaves use to do.
\end{itemize}

We observe that in both cases the system identifies the word ``black'' as toxic, but not ``white'', even when these non-toxic adjectives are the only difference between them. Furthermore, the system only identifies ``immigrants'' as toxic when appearing next to ``Mexican'' but not with ``American''.

This undesired discrimination happens because there are lots of racist comments in the dataset, which are obviously annotated as toxic. Given that it seems there are more comments against some specific ethnic groups than others, the system associates certain racial references with racism and thus with toxicity.

This is a problem that comes from the data, including the one used in the pre-training phase of BERT models. However, there are several de-bias techniques in the literature \cite{manzini2019black, sun2019mitigating, liang2020debiasing} that could be applied to our model to alleviate it.

\section{Conclusion}

In this work, we have presented a solution for the SemEval-2021 Task 5: Toxic Spans Detection competition, which is a challenging task due to the subjectivity of toxicity and the requirement of context knowledge.

During the development of our solution, a multi-depth DistilBERT model, we have proved the power of pre-trained models and transfer learning to a downstream task with limited data, at the same time that we have demonstrated the benefits of combining the outputs of multiple BERT models' layers for token classification.

With an F1-score of 68.54\% the presented model ranks 14 out of 91 participating teams in the competition and, although it presents some racial bias that could be corrected, from the qualitative results we conclude that it has a very good performance, hence being able to be used in real-life applications.

\section*{Acknowledgments}
We would like to thank the SemEval-2021 organizers for their effort in preparing the challenge and their help during the competition, as well as the reviewers for their constructive comments. We are also deeply grateful to Jordi Armengol-Estapé from Barcelona Supercomputing Center (BSC) for providing valuable insight into our project.

\bibliographystyle{acl_natbib}
\bibliography{acl2021}

\vfill\eject
\appendix
\section{Implementation Details}
\label{app:imp}

We have developed this project using PyTorch\footnote{https://pytorch.org/} and the Huggingface\footnote{https://huggingface.co/transformers/} implementation of transformers. Specifically, we have used BertModel and DistilBertModel pre-trained models. 

Although Huggingface includes models for token classification, the input dimension for their classification layer is exactly 768 (for BERT and DistilBERT), the dimension of the Transformer blocks' output. However, in our case, we are concatenating different outputs, so these dimensions will vary from one experiment to another. To mimic their token classification models, we manually add a Dropout layer and the final classification layer with the appropriate input dimension (in our case $768 \times \textrm{\#concat\_outputs}$).

To train the models we use Cross Entropy Loss with Label Smoothing. This type of regularization slightly changes the target vector, asking the model to predict $1-\epsilon$ for the correct class and $\epsilon$ for the others instead of the usual hard assignment of $1$ for the true class. As seen in Section~\ref{sec:ablation}, this technique helps the system improve as it helps modelling the intrinsic subjectivity of the data. For our best models we use $\epsilon = 0.1$.

We perform from 4 to 8 training epochs with Adam optimizer, learning rate $1e^{-5}$, batch size of 8 and 25\% dropout rate. Finally, we select the epoch with the best F1-score at trial set as our best checkpoint.

We keep the default values for the rest of parameters.

Each model has required approximately 15 minutes of training time on an NVIDIA Tesla V100 GPU. With the same hardware, the inference time is 430ms per sentence, so our system is able to work in Near Real Time (NRT).

\end{document}